\documentclass{article}

\usepackage[final]{neurips_2024}

\usepackage[utf8]{inputenc} 
\usepackage[T1]{fontenc}    
\usepackage{hyperref}       
\usepackage{url}            
\usepackage{booktabs}       
\usepackage{amsfonts}       
\usepackage{nicefrac}       
\usepackage{microtype}      
\usepackage{xcolor}         
\usepackage{amsmath}
\usepackage{multirow}

\title{SMART: Scalable Multi-agent Real-time Simulation via Next-token Prediction}

\author{%
  Wei~Wu\thanks{Equal contribution} \\
  Tsinghua University \\
  SenseTime Research \\
  \texttt{wuwei@senseauto.com} \\
  \And
  Xiaoxin~Feng\footnotemark[1] \\
  SenseTime Research \\
  \texttt{fengxiaoxin@senseauto.com} \\
  \AND
  Ziyan~Gao\footnotemark[1] \\
  SenseTime Research \\
  \texttt{gaoziyan@senseauto.com} \\
  \And
  Yuheng~Kan\\
  SenseTime Research \\
  \texttt{kanyuheng@senseauto.com} \\
}

\usepackage{graphicx}
\begin{document}

\maketitle

\begin{abstract}
  Data-driven autonomous driving motion generation tasks are frequently impacted by the limitations of dataset size and the domain gap between datasets, which precludes their extensive application in real-world scenarios. To address this issue, we introduce SMART, a novel autonomous driving motion generation paradigm that models vectorized map and agent trajectory data into discrete sequence tokens. These tokens are then processed through a decoder-only transformer architecture to train for the next token prediction task across spatial-temporal series. This GPT-style method allows the model to learn the motion distribution in real driving scenarios. SMART achieves state-of-the-art performance across most of the metrics on the generative Sim Agents challenge, ranking 1st on the leaderboards of Waymo Open Motion Dataset (WOMD), demonstrating remarkable inference speed. Moreover, SMART represents the generative model in the autonomous driving motion domain, exhibiting zero-shot generalization capabilities: Using only the NuPlan dataset for training and WOMD for validation, SMART achieved a competitive score of 0.72 on the Sim Agents challenge. Lastly, we have collected over 1 billion motion tokens from multiple datasets, validating the model's scalability. These results suggest that SMART has initially emulated two important properties: scalability and zero-shot generalization, and preliminarily meets the needs of large-scale real-time simulation applications. We have released all the code to promote the exploration of models for motion generation in the autonomous driving field. The source code is available at \url{https://github.com/rainmaker22/SMART}.
\end{abstract}

\section{Introduction}
In the context of autonomous driving, leveraging vectorized maps and vehicle trajectory data facilitates various motion generation tasks, including motion planning \cite{hu2023planning, cheng2023rethinking, huang2023gameformer, huang2023dtpp, cui2021lookout}, motion prediction \cite{wilson2023argoverse, ettinger2021large, song2020pip}, and Sim Agents \cite{gulino2024waymax}. Previous research \cite{cui2019multimodal, chen2023interaction, nayakanti2023wayformer} has predominantly employed encoder networks to represent driving scenes and decoder networks to generate multi-modal motions. These generated motions are then directly regressed to continuous trajectory distributions using Gaussian \cite{chai2019multipath} or Laplace \cite{zhou2023qcnext} mixture loss functions. While this framework demonstrates strong performance in prediction tasks that prioritize regression accuracy, it often underperforms in motion generative tasks that emphasize the safety and reasonableness of driving behavior, such as planning \cite{caesar2021NuPlan} or Sim Agents \cite{montali2024waymo}. The primary reasons for this underperformance are as follows: First, the framework does not represent future interactions between the motions of different agents, leading to inconsistent scene-level forecasting. Second, the model generates multi-modal motion by initializing multiple intention queries in the decoder, which is typically limited by GPU memory, resulting in a fixed number of motion modalities. Consequently, it is uncertain whether the generated modalities sufficiently represent the diversity of future behaviors. Thirdly, these models struggle to generalize across different datasets, requiring new data collection for training in new urban environments or maps. 

The advent of autoregressive large language models (LLMs) \cite{floridi2020gpt, touvron2023llama} has ushered in a new era in artificial intelligence. Drawing inspiration from this, some studies in the driving motion generation domain\cite{philion2023trajeglish, seff2023motionlm}, have tokenized agent trajectories into discrete motion tokens and employed a Next Token Prediction (NTP) task based on cross-entropy loss for autoregression. These models continue to utilize an encoder-decoder architecture, encoding continuous vectorized map and historical trajectory data with an encoder, and decoding discrete tokens solely in the decoder module. Compared to continuous distribution regression methods, the autoregressive paradigm of NTP has the following advantages: the model adopts a step-by-step next token prediction, allowing it to model interactions between agents' motions at each time step, and the number of modalities is not limited, leading to better diversity in generative tasks. 

However, existing NTP-based motion models still fail to address the aforementioned issues of generalizability and scalability, which have a critical impact on industrial applications. Generalizability means achieving satisfactory results across diverse datasets through zero-shot and few-shot learning, while scalability involves improving model performance as dataset size or model parameters increase, following scaling laws defined by \cite{hoffmann2022training}. This shortfall is due to two main factors: First, current model architectures lack generalizability under the constraints of limited data scale. Due to the high cost of acquiring extensive driving data, open-source datasets typically cover only a few hundred hours of driving in specific urban areas, with significant domain gaps caused by perceptual and regional differences. Second, unlike tasks involving the serialization of a single dimension, motion generation requires the serialization of both the temporal dimension of trajectories and the spatial interactions between maps and agents. To tackle these challenges, this paper introduces the SMART model: Scalable Multi-Agent Real-Time Motion Generation via Next-token Prediction. The model incorporates a tokenizer for map data and proposes an autoregressive prediction task for the next road token prediction to enhance the model's spatial comprehension. Subsequently, a GPT-style approach is adopted, tokenizing agent trajectories across the entire time series to establish a decoder-only transformer model. The decoder-only transformer allows SMART to compute the next token for the upcoming frame at the current moment during inference, eliminating the need to re-encode historical motion tokens with each inference, which significantly improves inference efficiency for real-time interactive autonomous driving simulation.

In summary, our contributions to the community include:
(1) We propose a novel framework for motion generation, incorporating a tokenization scheme for both vectorized road and agent trajectories and utilizing a decoder-only transformer for training on the next token prediction task. This approach offers new insights into the design of motion generation algorithms for autonomous driving. (2) In the field of driving motion generation, we have pioneered a focus on the model's zero-shot generalizability across different datasets. Notably, the model trained solely on the NuPlan dataset performed well on the WOMD test dataset, despite the lack of overlap between the map areas of these two datasets. An empirical validation of SMART models' scalability emulates the appealing properties of large fundamental models. (3) SMART achieves state-of-the-art performance across most metrics in the generative Sim Agents challenge, ranking \textbf{1\textsuperscript{th}} on the WOMD leaderboards\footnote{\url{https://waymo.com/open/challenges/2024/sim-agents/}}. Furthermore, SMART's single-frame inference time is within 15ms, meeting the real-time requirements for interactive simulation in autonomous driving.

\section{Related work}
\subsection{Properties of auto-regressive large models}
\paragraph{Scalability and zero-shot generalization} Power-law scaling laws \cite{kaplan2020scaling, floridi2020gpt, radford2019language} mathematically describe the relationship between the growth of model parameters, dataset sizes, computational resources, and the performance improvements of machine learning models, providing several distinct benefits. Firstly, they enable the extrapolation of a larger model's performance by scaling up model size, data size, and computational cost. Secondly, the scaling laws have demonstrated a consistent and non-saturating increase in performance, corroborating their sustained advantage in enhancing model capabilities. Zero-shot generation refers to the ability of models to generate predicted motions for time series from unseen datasets. Previous work \cite{orozco2020zero, jin2022domain} on zero-shot generation typically involves training on a single time series dataset and testing on a different dataset. In this study, we utilize the NuPlan dataset for training SMART models and the WOMD validation dataset for testing. Existing methods in the autonomous driving field \cite{sima2023drivelm, tian2024drivevlm} often rely on LLMs or VLMs to assist in decision-making and planning to enhance generalizability and interpretability. However, no studies have attempted to directly construct a foundational model for the driving motion field to validate scalability and zero-shot generalizability.

\subsection{Tokenizer in continuous domains} Language models \cite{touvron2023llama, touvron2023llama2} rely on Byte Pair Encoding or WordPiece algorithms for text tokenization. Visual generation models\cite{yu2024language, yu2022scaling} based on language models also necessitate the encoding of 2D images into 1D token sequences. Early endeavors VQVAE \cite{van2017neural} have demonstrated the ability to represent images as discrete tokens, although the reconstruction quality was relatively moderate. In the driving motion domain, MotionLM\cite{seff2023motionlm} used a simple uniform quantization of axis-aligned deltas between consecutive waypoints of agent trajectories. 

\subsection{Driving motion generation}
Our work builds heavily on recent advancements in driving motion generation. A comprehensive range of generative models has been applied to this problem, including continuous motion distribution regression \cite{salzmann2020trajectron++, amirloo2022latentformer, suo2021trafficsim}, diffusion models \cite{zhong2023guided, jiang2023motiondiffuser}, and discrete autoregressive models \cite{philion2023trajeglish, seff2023motionlm}. MotionDiffuser \cite{jiang2023motiondiffuser} is a diffusion-based representation method for modeling the joint distribution of future trajectories across multiple agents, leveraging a simple predictor design and PCA compression for efficient, top-performing multi-agent motion prediction. While these diffusion-based models produce multi-modal future trajectories of individual agents, they only capture the marginal distributions of possible agent movements and do not model interactions among agents' future motions. Typical distribution regression models use parametric continuous distributions such as Gaussian \cite{shi2023mtr++} or Laplace \cite{zhou2023qcnext} to model the future motion distribution. A limitation of these models is the uncertainty of whether the Gaussian or Laplace mixture distribution is flexible enough to represent the distribution over future states. Additionally, to generate multi-modal future motions, these models often need to incorporate motion goal candidates \cite{gu2021densetnt} or learnable latent embeddings \cite{varadarajan2022multipath++} as multi-modal queries in the decoder module, resulting in significant memory usage and increased inference time. MotionLM \cite{seff2023motionlm} treats multi-agent motion prediction in autonomous vehicles as a language modeling task, generating interactive trajectories through a simplified autoregressive process without requiring complex optimizations and latent anchor embeddings. On this basis, Trajeglish \cite{philion2023trajeglish} targets multi-agent offline closed-loop simulation.

\section{Method} \label{sec:3}
In this section, we introduce SMART, an autoregressive generative model for dynamic driving scenarios. While both language and agent motions are sequential, they differ in their representation—natural language consists of words from a finite vocabulary, whereas agent motions are continuous real-valued data. This distinction necessitates the unique design outlined in Sec.~\ref{sec:3.1} for agent motion and road vector tokenizer, including the construction of vocabulary and the tokenization of motion sequences. Sec.~\ref{sec:3.2} provides a comprehensive description of the model's architecture. Sec.~\ref{sec:3.3} elaborates on the training tasks designed for the proposed model to learn the distribution of the motion token within the temporal sequence and the distribution of the road token within the spatial sequence.

\subsection{Tokenization}\label{sec:3.1}
\begin{figure}[tb]
\centering
\includegraphics[width=1.0\linewidth]{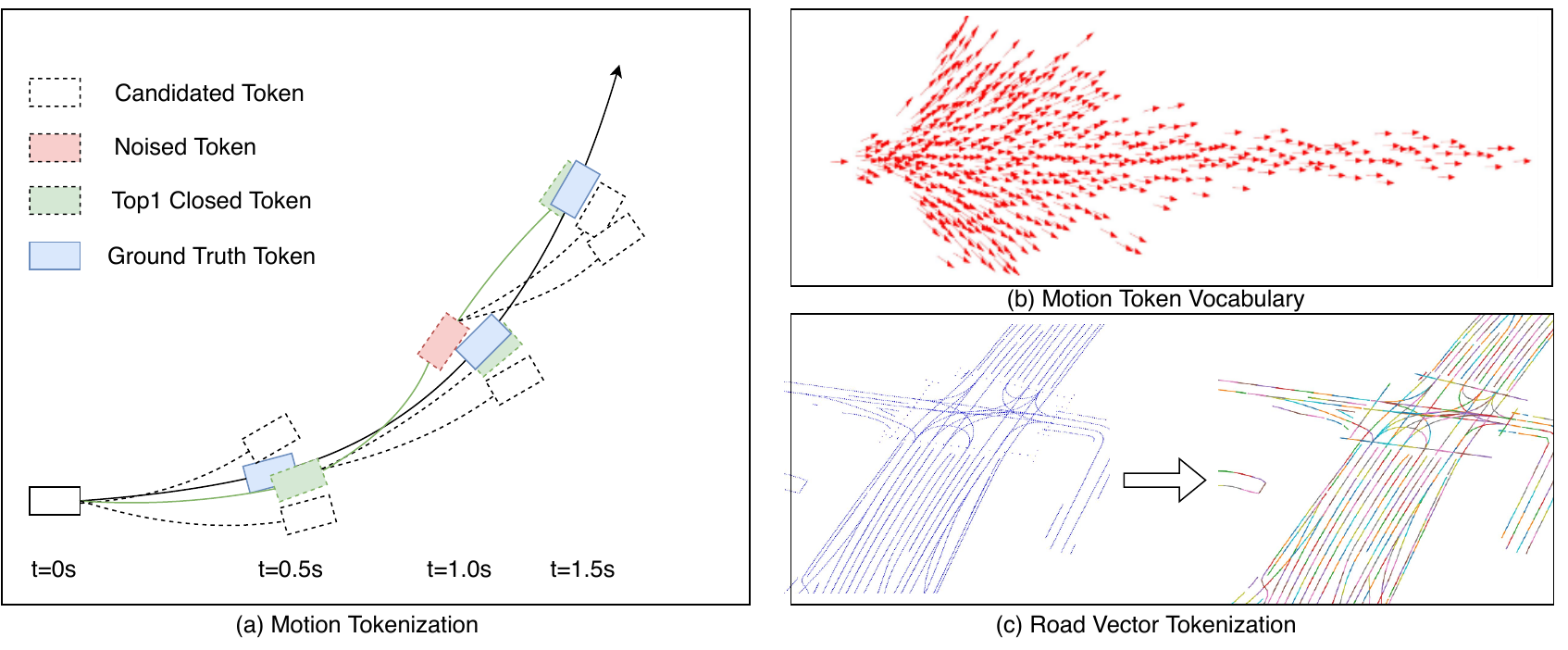}
\caption{(a) At time t=0s, the current vehicle state is used as the reference to select the token closest to the ground truth bounding box within the token set. At time t=0.5s, the matched token from the previous step is used to select the next predicted token. At time t=1.0s, a noised token serves as the reference to determine the token for t=1.5s. This iterative process continues. (b) Motion token vocabulary with time granularity equal to 0.5s.(c) The original road vector features are represented as continuous sequences of map points. We divide the original map into multiple segments, each within 5 meters in length, and then perform matching with discrete tokens. The final map is composed of road vector tokens represented by different colored segments.}
\label{fig:1}
\end{figure}
\paragraph{Agent Motion tokenization} To apply discrete sequence modeling in continuous domains, prior works typically follow one of two approaches: either use a pre-trained tokenizer, such as VQVAE \cite{van2017neural} or VQGAN \cite{esser2021taming}, to encode continuous features into discrete tokens, or normalize continuous features and divide continuous values into discrete slots at equal intervals \cite{ansari2024chronos, seff2023motionlm}. For the former approach, establishing a latent vocabulary often requires a large amount of raw data to train the tokenizer; otherwise, the tokenizer itself will be biased towards the pre-training dataset. Since our work aims to enable the model to generalize effectively when trained on a small number of data samples, SMART opts to discretize explicit trajectory and map features. Specifically, similar to \cite{philion2023trajeglish}, we segment the continuous trajectories of all agents in the dataset into trajectory sets by fixed time intervals $t=0.5s$. Then, we cluster the trajectory sets using the k-disks algorithm. As shown in Figure~\ref{fig:1}(b), the sampled trajectories serve as our final agent motion token vocabulary $V_{a}$. 

As shown in Figure~\ref{fig:1}(a), the blue box represents the tokens obtained after discretizing the ground truth trajectory. At every 0.5-second interval, a search is conducted within the token vocabulary for candidate tokens, from which an appropriate(closest) token is selected to represent the current moment. Note that to prevent matching errors that may occur during the tokenization process of the agent motion sequence, we implement a rolling matching approach for the entire continuous motion sentence in a given period $T$. This implies that the token for the next time step is matched by referring to the position of the token currently matched, rather than relying on the actual correct position. However, due to the transformer decoder must perform sequential inference step by step, this approach inevitably leads to out-of-distribution issues due to compounding errors\cite{rawte2023survey}. Especially, in the field of autonomous driving, these accumulated errors may result in collisions and off-map events\cite{9636795}. To address this issue, we introduce noise into the tokenization process to enable the model to simulate distribution shifts during training. Specifically, we perturb the currently matched token by selecting one from the top-k tokens closest to the ground truth token in the vocabulary. Then, in the next time step, we match the motion token based on the perturbed vehicle state. This data augmentation method allows the model to effectively handle issues such as distribution shifts and accumulated errors, thereby enhancing robustness in generative tasks. Finally, the agent motion token is represented as $A \in \mathbb{R}^{N_{A} \times N_{T} \times F_{A}}$, where $N_{A}$ denotes the total number of agents, and $N_T$ represents the number of time steps, with a feature size of $F_{A}$, containing coordinates, heading, and shapes.
\paragraph{Road vector tokenization}
To enhance the model's generalization capabilities, we have applied a similar tokenization process to road vectors as we did with agent motion. Each road vector is a directed lane segment with features including start and end positions, length, turn direction, and other semantics from the dataset. To obtain fine-grained inputs for the road network, all road vectors are segmented into tokens spanning no longer than 5 meters in length. Unlike the motion sequence, the tokenization process of the road sentence does not have a time-series dependency. As shown in Figure~\ref{fig:1}(c), the tokenization of the road sentence is performed in parallel, directly tokenizing all the original road vector segments. The road vector token is represented as $R \in \mathbb{R}^{N_{R} \times F_{R}}$, where $N_{R}$ denotes the total number of road vectors, and $F_{A}$ represents the token features.

\subsection{Model Architecture} \label{sec:3.2}
\begin{figure}[tb]
\centering
\includegraphics[width=0.9\linewidth]{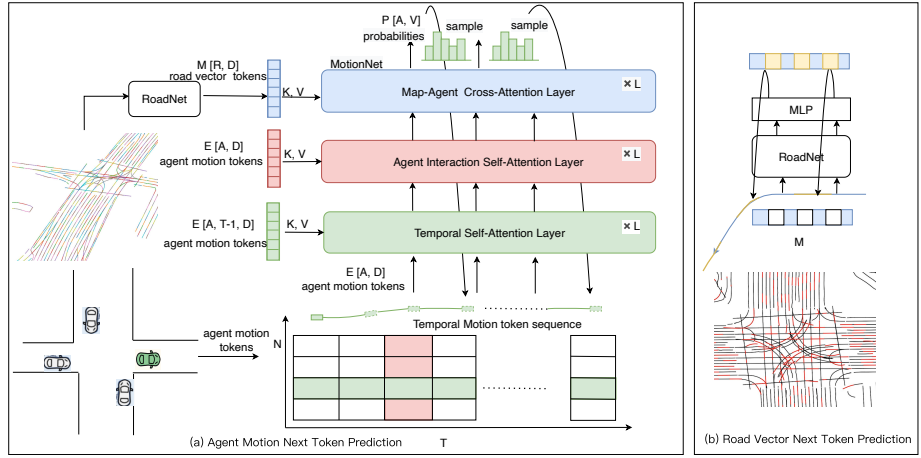}
\caption{The architecture of SMART framework (a) We train a decoder-only transformer that predicts the motion tokens of multi-agents conditional on previous motion tokens, interactive agent motion tokens, and encoding road tokens. The model is trained to predict the next motion token. (b) Illustration for our proposed road spatial understanding training task.}
\label{fig:2}
\end{figure}

Figure~\ref{fig:2} illustrates the simple but expressive model architecture of SMART. The model comprises an encoder for road map encoding and a motion decoder that predicts a category distribution based on motion token embeddings.

\paragraph{RoadNet: road token encoder}
We employ multi-head self-attention (MHSA) to model the relationships among road tokens, after which the updated road token encodings will assist motion token decoding. For the $i^{th}$ road token, we derive a query from its embedding $r_{i}$ and let it attend to the neighboring tokens ${r_{j}} \in R_{i}$:
\begin{equation}\label{1}
\begin{aligned}
r_{i'} = &\text{MHSA}\left(q(r_{i}),\right. \left.k(r_{j}, \text{RPE}_{ij}), v(r_{j}, \text{RPE}_{ij}\right), \quad j \in R_{i}
\end{aligned}
\end{equation}
where $R_{i}$ denotes the neighbor set of the road tokens. To incorporate spatial awareness for map encoding, we generate the $j^{th}$ key/value vector from the concatenation of $r_{j}$ and the relative positional embedding $\text{RPE}_{ij}$\cite{cui2023gorela}.

\paragraph{MotionNet: factorized agent motion decoder}
Prevailing methods for encoding agents prioritize capturing the temporal dynamics of an agent's movements, followed by the integration of agent-map and agent-agent interactions, as highlighted by \cite{shi2022motion}. Factorized attention effectively captures detailed agent-map interactions across temporal scales \cite{ngiam2021scene}. In our work, we leverage a factorized Transformer architecture with multi-head cross-attention (MHCA) to decode complex road-agent and agent-agent relationships along the time series. Akin to query-centric methodologies \cite{zhou2023query}, we utilize relative positional embeddings to differentiate between agents' local coordinate frames, enabling symmetric encoding. Take the $i^{th}$ agent at time step $t$ as an example. Denoted as Eq.\ref{eq:2:a}, given the query derived from the agent motion token's embedding $e_{i}^{t}$, we employ temporal attention by computing the key and value based on which are the $i^{th}$ agent's token embeddings from time step $t-\tau$ to time step $t-1$ and the corresponding relative positional embeddings. 

\begin{subequations}\label{eq:2}
\begin{equation}\label{eq:2:a}
\begin{split}
e_{i'} = \text{MHSA}\left(q(e_{i}^{t}), k(e_{i}^{t-\tau}, \text{RPE}_{i}^{t, t-\tau}), v(e_{i}^{t-\tau}, \text{RPE}_{i}^{t, t-\tau})\right), \quad 0<\tau<t
\end{split}
\end{equation}
\begin{equation}\label{eq:2:b}
e_{i'} = \text{MHCA}\left(q(e_{i}^{t}), k(r_{j}, \text{RPE}_{ij}), v(r_{j}, \text{RPE}_{ij})\right), \quad j \in N_{i}
\end{equation}
\begin{equation}\label{eq:2:c}
e_{i'} = \text{MHSA}\left(q(e_{i}^{t}), k(e_{j}^{t}, \text{RPE}_{ij}^{t}), v(e_{j}^{t}, \text{RPE}_{ij}^{t})\right), \quad j \in N_{i}
\end{equation}
\end{subequations}

Likewise, in Eq.\ref{eq:2:b} and Eq.\ref{eq:2:c}, the key and value for agent-map and agent-agent attention are derived from road token $r_{j}, j \in N_{i}$ and agents' motion token $e_{j}^{t}, j \in N_{i}$ in the neighborhood respectively, where the neighbor set $N_{i}$ is determined by a distance threshold of 50 meters. We stack the temporal, the agent-agent, and the agent-map attention sequentially as one fusion block and repeat such blocks $K$ times.

\subsection{Spatial-temporal next token prediction} \label{sec:3.3}
In the training stage, we train SMART to understand the temporal and spatial relationships in the traffic scene. This is achieved with two next token prediction tasks on RoadNet and MotionNet, the model is optimized with the summation of the two tasks' objectives.

\paragraph{Road vector next token prediction}
As shown in Figure~\ref{fig:2}(b), the road vector NTP task targets RoadNet to learn the spatial structure of road vector inputs. Unlike agent motions, road vectors form a graph rather than a sequence, making it challenging to apply next token prediction tasks directly. To address this issue, we extract the original topological information of roads and model the road vector tokens with sequential relationships based on their predecessor-successor connections. As depicted in Figure~\ref{fig:2}(b), in the pre-training NTP task, the subsequent road vector token is predicted using the preceding road token based on the road topology. This approach requires RoadNet to understand the connectivity and continuity among unordered road vectors. The loss function for a single tokenized road vector sequence is defined as:

\begin{equation}
\begin{aligned}
\text{loss}(\gamma) = -\sum_{j=1}^{J}\sum_{i=1}^{V_{r}} (\text{r}_{i}^{j+1}==\text{r}_{i^{gt}}^{j+1}) \log(p_{\gamma}(\text{r}_{i}^{j+1}|\text{r}_{i}^{1:j
}))
\end{aligned}
\end{equation}

where $p_{\gamma}(\text{r}_{i}^{j+1}|\text{r}_{i}^{1:j
})$ denotes the categorical distribution predicted by the RoadNet parameterized by $\gamma$, $J$ represents a complete polyline that has not yet been split into road vector tokens, $\text{r}^{1:j}$ Representing the road token embedding of the predecessor, and $\text{r}_{i}^{j+1}$ is the next predicted road vector token. This loss function ensures that RoadNet learns to predict the correct next road vector token given the preceding tokens, thereby capturing the spatial continuity and connectivity within the road network. 

\paragraph{Motion next token prediction}
Motion NTP task targets MotionNet to understand not only the temporal dependencies in agents' motions but also the spatial dependencies between agent-map and agent-agent. SMART is trained to minimize the cross entropy between the distribution of the ground truth token label and the predicted distribution. Formally, the loss function for a single tokenized motion sentence is given by:
\begin{equation}\label{3}
\begin{aligned}
loss(\theta) = -\sum_{t=1}^{T}\sum_{i=1}^{V_{a}} (a_{i}^{t+1}==a_{i^{gt}}^{t+1}) log(p_{\theta}(a_{i}^{t+1}|e_{i}^{1:t}, r_{j}))
\end{aligned}
\end{equation}
where $p_{\theta}(a_{i}^{t+1}|e_{i}^{1:t}, m_{j})$ denotes the categorical distribution predicted by the model parameterized by $\theta$, $e_{1:t}$ is the historical tokenized agent motion embeddings, $a_{i}^{t+1} \in A$ is the next predicted agent motion token and $ r_{j}$ is the tokenized nearby road vector series. Note that SMART performs autoregression via classification\cite{torgo1996regression}. Opting for a categorical output distribution offers a key advantage: it imposes no restrictions on the structure of the output distribution, allowing the model to learn arbitrary distributions, including multimodal ones. This flexibility is especially valuable for a fundamental model, as agent and road tokens from diverse datasets may follow distinct output distribution patterns.

\section{Experiments} \label{sec:4}
\label{headings}
To validate the generalizability and scalability of the SMART model, we conducted extensive experiments and trained models across various scales. On the official WOMD Sim Agents Challenge (WOSAC), we employed the SMART 7 Million parameters (7M) model, which was exclusively trained on the WOMD dataset. Concurrently, the SMART 7M model was also utilized for generalization experiments and ablation studies. In the scale law experiments, we integrated additional datasets and trained on models of multiple scales. For all experiments, the testing datasets employed the split validation dataset from WOMD. Detailed hyperparameters for the SMART architecture can be found in Section \ref{sec:5:2}. In the following sections, Section \ref{sec:4.1} presents the results of rollouts generated by SMART on the WOSAC benchmark \cite{montali2024waymo}. Evaluations of SMART's generalizability and scalability are detailed in Sections \ref{sec:4.2} and \ref{sec:4.3}, respectively. Finally, an ablation analysis of our design methods is conducted in Section \ref{sec:4.4}.

\subsection{Comparison for motion generation task} \label{sec:4.1}
\paragraph{Performance comparison}
We compare proposed SMART with existing motion generation approaches including diffusion models\cite{guo2023scenedm}, continuous distribution regression models \cite{wang2023multiverse, shi2023mtr++}, and next token autoregressive model\cite{philion2023trajeglish}. Because the Sim Agents challenge metrics were changed twice, to compare it more broadly with the previous methodology, we test the performance of our model using both the WOMD Sim Agents 2023 and 2024 Benchmark\cite{montali2024waymo}. As shown in Table~\ref{tab:1} and Table~\ref{tab:2}, SMART achieves not only the best Realism Meta metric but also a high prediction precision. SMART's modeling approach for maps and motion enables it to learn the behavioral distribution within the data more effectively than prior work. Notably, SMART-zeroshot represents a model trained solely on the NuPlan dataset and directly inferred on the Waymo test set. As shown in Table~\ref{tab:2}, it achieves performance close to that of MVTE. For further detailed comparisons, please refer to \ref{sec:5:1}.

\begin{table}[t]
    \caption{Comparison with state-of-the-art models on WOMD 2023 Sim Agents benchmark}
    \centering
    \begin{tabular}{cccccc}
         \toprule
         Method&  \begin{tabular}{c}Realism \\Meta metric$\uparrow$\end{tabular}&  \begin{tabular}{c}Kinematic \\ metrics$\uparrow$\end{tabular}& \begin{tabular}{c} Interactive \\metrics$\uparrow$\end{tabular} & \begin{tabular}{c} Map-based \\ metrics$\uparrow$\end{tabular} & minADE $\downarrow$\\
         \midrule
         SMART 7M&  \textbf{0.6587}&  0.4190&  \textbf{0.8014}&  \textbf{0.8523}&  1.7453\\
         Trajeglish\cite{philion2023trajeglish}&  0.6451&  0.4166&  0.7845&  0.8216&  1.5712\\
         MVTE\cite{wang2023multiverse}&  0.6448&  0.4202&  0.7666&  0.8387&  1.6770\\
         VPD-PRIOR&  0.6315&  \textbf{0.4261}&  0.7233&  0.8330&  1.3400\\
 QCNeXt\cite{zhou2023qcnext}& 0.4538& 0.3109& 0.5654& 0.5051& \textbf{1.0830}\\ MultiPath\cite{varadarajan2022multipath++}& 0.4766& 0.1792& 0.6380& 0.6866&2.0517\\
    \bottomrule
    \end{tabular}
    \label{tab:1}
\end{table}
\begin{table}[ht]
    \caption{Comparison with state-of-the-art models on WOMD 2024 Sim Agents benchmark}
    \centering
    \setlength{\tabcolsep}{5pt}
    \begin{tabular}{cccccc}
    \toprule
         Method&  \begin{tabular}{c}Realism \\Meta metric$\uparrow$\end{tabular}&  \begin{tabular}{c}Kinematic \\ metrics$\uparrow$\end{tabular}&  \begin{tabular}{c} Interactive \\metrics$\uparrow$\end{tabular} &  \begin{tabular}{c} Map-based \\ metrics$\uparrow$\end{tabular} &   minADE $\downarrow$\\
    \midrule
    SMART 101M & \textbf{0.7614}& \textbf{0.4786}& \textbf{0.8066}& \textbf{0.8648}&\textbf{1.3728}\\
    SMART 7M & 0.7591& 0.4759& 0.8039&0.8632&1.4062\\
    BehaviorGPT&  0.7473& 0.4333&  0.7997&  0.8593&  1.4147\\
    GUMP&  0.7431&  0.4780&  0.7887&  0.8359&  1.6041\\
    MVTE& 0.7302& 0.4503& 0.7706& 0.8381&1.6770\\
    SMART-zeroshot& 0.7210& 0.4311& 0.7806& 0.8099&2.5703\\
    VBD& 0.7200& 0.4169& 0.7819& 0.8137&1.4743\\
    TrafficBOTv1.5& 0.6988& 0.4304& 0.7114& 0.8360&1.8825\\
    congniBOTv1.5& 0.6288& 0.3293& 0.7129& 0.6918&-\\
    \bottomrule
    \end{tabular}
    \label{tab:2}
\end{table}

\paragraph{Efficiency comparison}
SMART also demonstrates remarkable speed in multi-agent motion generation. Previous encoder-decoder models \cite{seff2023motionlm, shi2023mtr++} suffer from high computational costs, as the model requires multiple query embeddings in the decoder module to generate multi-modal motions. Benefiting from the advantages of the decoder-only transformer architecture, SMART only needs to compute the next token for the upcoming frame at the current moment during inference, without the need to re-encode historical motion tokens. By reusing the token embeddings computed in previous observation time horizons, the complexity of the agent motion decoder is reduced to $O(N_{A}N_{T}) + O(N_{A}N_{R}) + O(N_{A}^2)$. In contrast, for encoder-decoder models like \cite{makansi2019overcoming}, besides the computational load of the encoder module, additional computations of $O(N_{A}^{2}N_{M})+O(N_{A}N_{M}N_{R})$ are required for generating multi-modalities of trajectories, where $N_{M}$ represents the number of modalities. The average single-step inference time of SMART is influenced by the number of map tokens and agent motion tokens, fluctuating between 5 to 20 ms, and averaging under 10 ms. Thus, it significantly meets the current needs of interactive real-time online simulation in autonomous driving.

\subsection{Generalization} \label{sec:4.2}
\paragraph{Zero-shot generalization on different dataset}
Zero-shot generation is the ability of models to generate motions for time series from different datasets. In this work, we use the training data from NuPlan dataset to train SMART models and the test data from WOMD validation dataset.
As shown in Table~\ref{tab:3}, SMART\textsuperscript{*} still achieves good performance in the overall metrics. Due to significant differences in the accuracy of the calibrated ground truth values for agent position and heading between different datasets, there may be a larger gap in the agent kinematic metrics, resulting in lower scores. However, SMART\textsuperscript{*} demonstrated excellent generalization in the metrics of agent interaction and drivable map. It is worth mentioning that the size of the two datasets does not differ greatly, so the SMART model can have good generalization ability based on a small number of data training. 
\begin{table}[ht]
    \caption{Zero-shot generalization on different datasets. SMART denotes a model trained on WOMD only. SMART\textsuperscript{*} denotes a model trained on NuPlan dataset only. SMART\textsuperscript{**} denotes a model after 1 epoch of finetuning with an initial learning rate of 0.0001 on WOMD based on SMART\textsuperscript{*} model.}
    \centering
    \begin{tabular}{ccccc}
    \toprule
         Method&  \begin{tabular}{c}Kinematic \\ metrics$\uparrow$\end{tabular}&  \begin{tabular}{c} Interactive \\metrics$\uparrow$\end{tabular} &  \begin{tabular}{c} Map-based \\ metrics$\uparrow$\end{tabular} &   minADE $\downarrow$\\
    \midrule
    SMART& 0.4537& 0.8034& 0.8514&1.5127\\
    SMART\textsuperscript{*} & 0.4161& 0.7853& 0.7970& 2.3041\\
    SMART\textsuperscript{**} & 0.4310&  0.8087&  0.8559&  1.5671\\
    \bottomrule
    \end{tabular}
    \label{tab:3}
\end{table}

\begin{figure}[tb]
\centering
\includegraphics[width=1.0\linewidth]{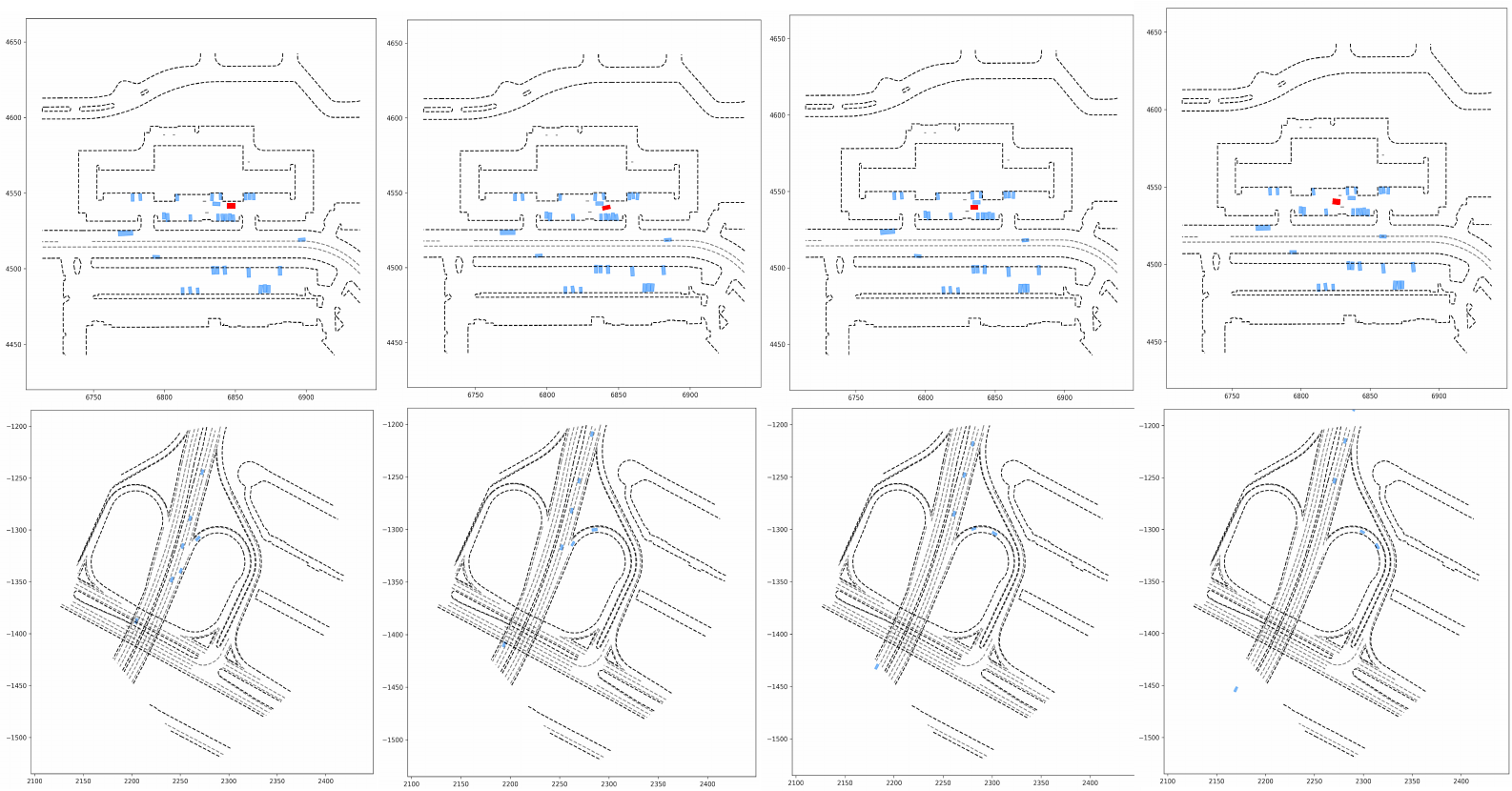}
\caption{Qualitative results of closed-loop planning for two representative scenarios from the test set. Each scenario (every row) lasts 8 seconds and we take 4 snapshots with a 2-second interval. SMART controls all the agents in the scenario. The first row depicts a parking lot area. The red vehicle in the picture effectively completed a detour around a stationary vehicle ahead in the parking lot. The second row shows a scene of a large curvature U-turn in a ramp zone, where the traffic flow in the right lane of the ramp has completed the behavior of ramp exit under the control of SMART. It is recommended to refer to supplementary materials for more videos}
\label{fig:3}
\end{figure}

\paragraph{Zero-shot generalization on unseen scenarios} 
Multiple map scenarios as shown in Figure~ \ref{fig:3} are present only in the WOMD but not in the NuPlan dataset. Without modifications to the network architecture or tuning parameters, SMART trained only on NuPlan has achieved decent results in these scenarios, substantiating the generalization ability of SMART.

\subsection{Scalability} \label{sec:4.3}
Prior research \cite{kaplan2020scaling, touvron2023llama} have established that scaling up large language models (LLMs) leads to a predictable decrease in test loss $L$. This trend correlates with parameter counts $N$, training tokens $T$, following a power-law: 
\begin{equation}\label{5}
\begin{aligned}
log(L) = \beta log(X) + \alpha
\end{aligned}
\end{equation}
where $X$ can be any of $N$, $T$. The exponent $\alpha$ reflects the smoothness of power-law, and $L$ denotes the reducible loss normalized by irreducible loss. The data sources for validating scaling laws are detailed in the \ref{sec:5:3}. Overall, we trained models across four sizes, ranging from 1M to 100M parameters, on a training set containing 2.2M scenarios (or 1B motion tokens under 0.5s agent motion tokenization).
\begin{figure}[htb]
\centering
\includegraphics[width=0.8\linewidth]{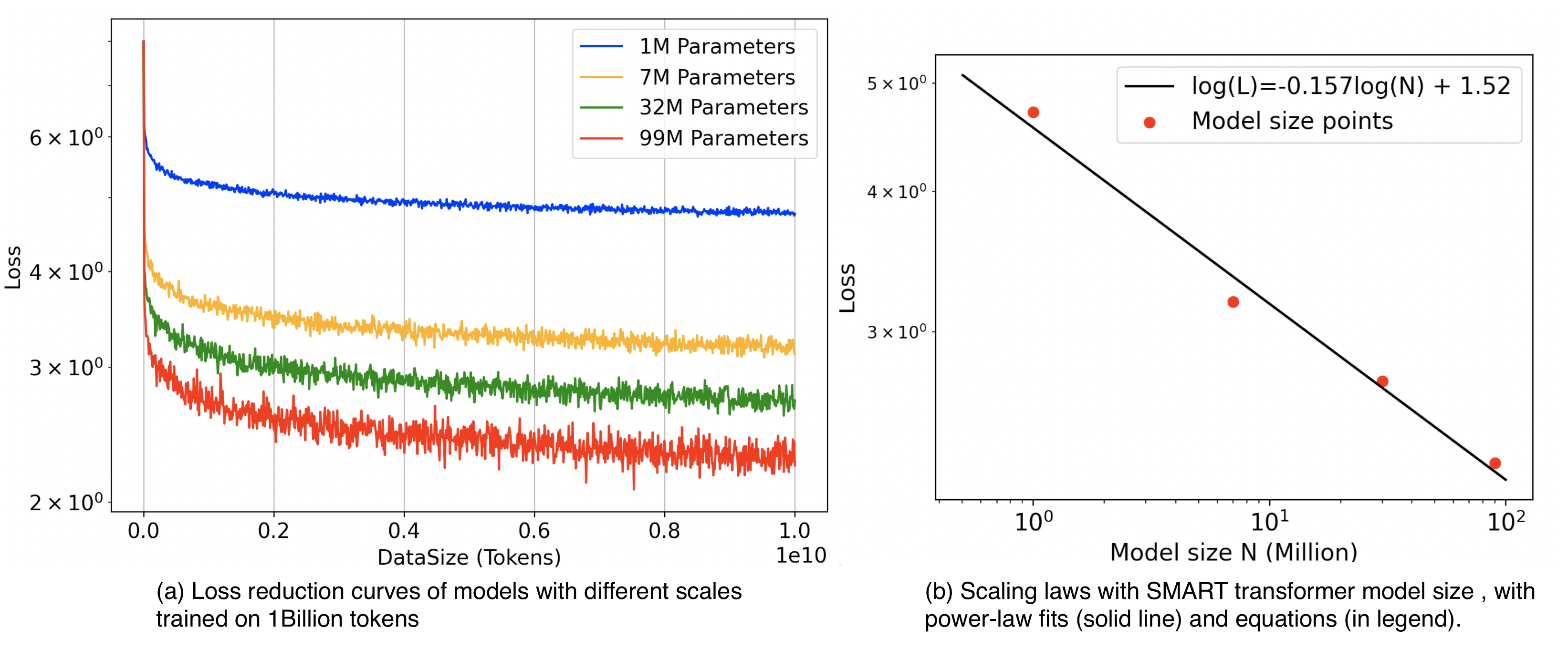}
\caption{Due to limitations in dataset size, we trained models at multiple scales ranging from 1M to 101M on a total of 1 billion tokens. (a) Training loss of different models (b) Axes are all on a logarithmic scale. The power-law scaling law can be expressed as a solid line. Exponents $\beta=-0.157$ suggest a smooth decline in test loss $L$ when scaling up SMART models.}
\label{fig:4}
\end{figure}

\paragraph{Scaling laws with model parameters}
We investigate the test loss trend as the model size increases. We assessed the final test cross-entropy loss $L$ on the validation set of 100,000 traffic scenarios. The results are plotted in Figure~\ref{fig:4}, where we observed a clear power-law scaling trend for Loss $L$ as a function of model size $N$. The power-law scaling laws can be expressed as:
\begin{equation}\label{6}
\begin{aligned}
log(L) = -0.157log(X)+1.52
\end{aligned}
\end{equation}
These results verify the strong scalability of SMART, providing valuable insights into how model performance scales with dataset size.

\subsection{Ablation} \label{sec:4.4}
In this study, we aim to verify the effectiveness of each component of SMART. Results are reported in Table~\ref{tab:4}. The initial model, denoted as $M1$, is constructed on the architecture depicted in Sec.3.2, employing solely agent tokenization. The introduction of the road vector tokenization in $M2$, which tokenized the road vector states into discrete tokens, results in marked improvements over $M1$ in the generalization capability. Comparing models $M1$ and $M2$ reveals that when trained solely on the WOMD dataset, the tokenization of road vectors results in a certain reduction in overall metrics. We speculate that discretized map tokens may lose some fine-grained geometric information about roads. $M4$ incorporates noised agent motion tokenization, designed to address cumulative errors and distributional shifts during inference. This modification leads to enhancements in both the interaction metric and the map-based metric.

\begin{table*}[t]
\caption{Ablation study on each component of SMART. Experimental results are based on the \textbf{WOMD validation set}. "RVT" indicates road vector tokenization, "RVNTP" indicates road vector next token prediction, "NAT" indicates noised agent tokenization, "NRVT" indicates noised road vector tokenization}
\centering
\scalebox{0.9}{
\setlength{\tabcolsep}{2pt} 
\begin{tabular}{@{}ccccccccccc} 
\hline
    SMART&&& && \multicolumn{3}{c}{Train on WOMD}& \multicolumn{3}{c}{Train on NuPlan}\\
    \hline
    \begin{tabular}{c}Model\\Number\end{tabular}& \begin{tabular}{c} RVT\end{tabular} & \begin{tabular}{c}  NAT\end{tabular}&\begin{tabular}{c} NRVT\end{tabular} &\begin{tabular}{c} RVNTP\end{tabular}& \begin{tabular}{c}kinematics\end{tabular} & \begin{tabular}{c}interactive\end{tabular}& \begin{tabular}{c}map\end{tabular}& \begin{tabular}{c}kinematics\end{tabular} & \begin{tabular}{c}interactive\end{tabular}&\begin{tabular}{c}map\end{tabular}\\
    \hline
    $M_{1}$&  &  & && \textbf{0.459} & \textbf{0.827}& \textbf{0.857}& 0.376& 0.593& 0.603\\
 $M_{2}$& $\surd$& &  && 0.434& 0.807& 0.840& 0.389& 0.696&0.724\\
 $M_{3}$& $\surd$ & $\surd$&  && 0.448& 0.809& 0.848& 0.413& 0.750&0.743\\
 $M_{4}$& $\surd$ & $\surd$& $\surd$ && 0.437& 0.801& 0.837& 0.411& 0.747&0.741\\
 $M_{5}$& $\surd$ & $\surd$ & & $\surd$ & 0.453& 0.813& 0.853& 0.413& 0.780&0.785\\
 $M_{6}$& $\surd$  & $\surd$ & $\surd$ &$\surd$ & 0.453& 0.803& 0.851& \textbf{0.416}& \textbf{0.785}& \textbf{0.797}\\
\hline
\end{tabular}
}
\label{tab:4}
\end{table*}

\section{Conclusions} \label{sec:6}
In this paper, we have introduced SMART, a novel paradigm for autonomous driving motion generation that leverages vectorized map and agent trajectory data, processed through a decoder-only transformer architecture in a GPT-style framework. We have observed that SMART emulates two critical properties: scalability and zero-shot generalization, which are essential for advancing large models. We believe that our findings and the release of all codes will encourage further exploration and development of models for motion generation in the autonomous driving field, ultimately contributing to more reliable autonomous driving systems.
\paragraph{Limitations} 
In this work, we primarily focus on the design of the learning paradigm and maintain a relatively simple design for the discrete token vocabulary. We believe that iterating SMART with an advanced tokenizer\cite{mentzer2023finite} or sampling technique can further improve the performance. Although we have collected training data from multiple datasets, we are still limited by the dataset size when validating the model's scalability, restricting us to models with a maximum scale of 100 million parameters. Given the focus of this work on generalization and scaling laws, a large number of hyperparameter ablation experiments remain to be verified, including the time granularity of agent motion tokens and the size of the token vocabulary. As a motion generation model, the ability of SMART to migrate to planning and prediction tasks still needs to be verified, and this is our top priority for future work.
\paragraph{Acknowledgement}
We thank the anonymous reviewers, area chairs, and program committee members for their valuable suggestions, which have greatly improved the quality of our work. We also appreciate the thoughtful discussions with Yue Gong and Shuxiang Lu. Authors Wei Wu, Xiaoxin, and Ziyan contributed equally to this work. Wei Wu led the project and provided funding support, while Xiaoxin and Ziyan focused on algorithm design, implementation, model training, and manuscript writing.

\bibliographystyle{plain}
\bibliography{main}

\begin{thebibliography}{10}

\bibitem{amirloo2022latentformer}
Elmira Amirloo, Amir Rasouli, Peter Lakner, Mohsen Rohani, and Jun Luo.
\newblock Latentformer: Multi-agent transformer-based interaction modeling and trajectory prediction.
\newblock {\em arXiv preprint arXiv:2203.01880}, 2022.

\bibitem{ansari2024chronos}
Abdul~Fatir Ansari, Lorenzo Stella, Caner Turkmen, Xiyuan Zhang, Pedro Mercado, Huibin Shen, Oleksandr Shchur, Syama~Sundar Rangapuram, Sebastian~Pineda Arango, Shubham Kapoor, et~al.
\newblock Chronos: Learning the language of time series.
\newblock {\em arXiv preprint arXiv:2403.07815}, 2024.

\bibitem{caesar2021NuPlan}
Holger Caesar, Juraj Kabzan, Kok~Seang Tan, Whye~Kit Fong, Eric Wolff, Alex Lang, Luke Fletcher, Oscar Beijbom, and Sammy Omari.
\newblock nuplan: A closed-loop ml-based planning benchmark for autonomous vehicles.
\newblock {\em arXiv preprint arXiv:2106.11810}, 2021.

\bibitem{chai2019multipath}
Yuning Chai, Benjamin Sapp, Mayank Bansal, and Dragomir Anguelov.
\newblock Multipath: Multiple probabilistic anchor trajectory hypotheses for behavior prediction.
\newblock {\em arXiv preprint arXiv:1910.05449}, 2019.

\bibitem{chen2023interaction}
Yongli Chen, Shen Li, Xiaolin Tang, Kai Yang, Dongpu Cao, and Xianke Lin.
\newblock Interaction-aware decision making for autonomous vehicles.
\newblock {\em IEEE Transactions on Transportation Electrification}, 2023.

\bibitem{cheng2023rethinking}
Jie Cheng, Yingbing Chen, Xiaodong Mei, Bowen Yang, Bo~Li, and Ming Liu.
\newblock Rethinking imitation-based planner for autonomous driving.
\newblock {\em arXiv preprint arXiv:2309.10443}, 2023.

\bibitem{cui2021lookout}
Alexander Cui, Sergio Casas, Abbas Sadat, Renjie Liao, and Raquel Urtasun.
\newblock Lookout: Diverse multi-future prediction and planning for self-driving.
\newblock In {\em Proceedings of the IEEE/CVF International Conference on Computer Vision}, pages 16107--16116, 2021.

\bibitem{cui2023gorela}
Alexander Cui, Sergio Casas, Kelvin Wong, Simon Suo, and Raquel Urtasun.
\newblock Gorela: Go relative for viewpoint-invariant motion forecasting.
\newblock In {\em 2023 IEEE International Conference on Robotics and Automation (ICRA)}, pages 7801--7807. IEEE, 2023.

\bibitem{cui2019multimodal}
Henggang Cui, Vladan Radosavljevic, Fang-Chieh Chou, Tsung-Han Lin, Thi Nguyen, Tzu-Kuo Huang, Jeff Schneider, and Nemanja Djuric.
\newblock Multimodal trajectory predictions for autonomous driving using deep convolutional networks.
\newblock In {\em 2019 International Conference on Robotics and Automation (ICRA)}, pages 2090--2096. IEEE, 2019.

\bibitem{esser2021taming}
Patrick Esser, Robin Rombach, and Bjorn Ommer.
\newblock Taming transformers for high-resolution image synthesis.
\newblock In {\em Proceedings of the IEEE/CVF conference on computer vision and pattern recognition}, pages 12873--12883, 2021.

\bibitem{ettinger2021large}
Scott Ettinger, Shuyang Cheng, Benjamin Caine, Chenxi Liu, Hang Zhao, Sabeek Pradhan, Yuning Chai, Ben Sapp, Charles~R Qi, Yin Zhou, et~al.
\newblock Large scale interactive motion forecasting for autonomous driving: The waymo open motion dataset.
\newblock In {\em Proceedings of the IEEE/CVF International Conference on Computer Vision}, pages 9710--9719, 2021.

\bibitem{floridi2020gpt}
Luciano Floridi and Massimo Chiriatti.
\newblock Gpt-3: Its nature, scope, limits, and consequences.
\newblock {\em Minds and Machines}, 30:681--694, 2020.

\bibitem{gu2021densetnt}
Junru Gu, Chen Sun, and Hang Zhao.
\newblock Densetnt: End-to-end trajectory prediction from dense goal sets.
\newblock In {\em Proceedings of the IEEE/CVF International Conference on Computer Vision}, pages 15303--15312, 2021.

\bibitem{gulino2024waymax}
Cole Gulino, Justin Fu, Wenjie Luo, George Tucker, Eli Bronstein, Yiren Lu, Jean Harb, Xinlei Pan, Yan Wang, Xiangyu Chen, et~al.
\newblock Waymax: An accelerated, data-driven simulator for large-scale autonomous driving research.
\newblock {\em Advances in Neural Information Processing Systems}, 36, 2024.

\bibitem{guo2023scenedm}
Zhiming Guo, Xing Gao, Jianlan Zhou, Xinyu Cai, and Botian Shi.
\newblock Scenedm: Scene-level multi-agent trajectory generation with consistent diffusion models.
\newblock {\em arXiv preprint arXiv:2311.15736}, 2023.

\bibitem{hoffmann2022training}
Jordan Hoffmann, Sebastian Borgeaud, Arthur Mensch, Elena Buchatskaya, Trevor Cai, Eliza Rutherford, Diego de~Las Casas, Lisa~Anne Hendricks, Johannes Welbl, Aidan Clark, et~al.
\newblock Training compute-optimal large language models.
\newblock {\em arXiv preprint arXiv:2203.15556}, 2022.

\bibitem{hu2023planning}
Yihan Hu, Jiazhi Yang, Li~Chen, Keyu Li, Chonghao Sima, Xizhou Zhu, Siqi Chai, Senyao Du, Tianwei Lin, Wenhai Wang, et~al.
\newblock Planning-oriented autonomous driving.
\newblock In {\em Proceedings of the IEEE/CVF Conference on Computer Vision and Pattern Recognition}, pages 17853--17862, 2023.

\bibitem{huang2023dtpp}
Zhiyu Huang, Peter Karkus, Boris Ivanovic, Yuxiao Chen, Marco Pavone, and Chen Lv.
\newblock Dtpp: Differentiable joint conditional prediction and cost evaluation for tree policy planning in autonomous driving.
\newblock {\em arXiv preprint arXiv:2310.05885}, 2023.

\bibitem{huang2023gameformer}
Zhiyu Huang, Haochen Liu, and Chen Lv.
\newblock Gameformer: Game-theoretic modeling and learning of transformer-based interactive prediction and planning for autonomous driving.
\newblock {\em arXiv preprint arXiv:2303.05760}, 2023.

\bibitem{jiang2023motiondiffuser}
Chiyu Jiang, Andre Cornman, Cheolho Park, Benjamin Sapp, Yin Zhou, Dragomir Anguelov, et~al.
\newblock Motiondiffuser: Controllable multi-agent motion prediction using diffusion.
\newblock In {\em Proceedings of the IEEE/CVF Conference on Computer Vision and Pattern Recognition}, pages 9644--9653, 2023.

\bibitem{jin2022domain}
Xiaoyong Jin, Youngsuk Park, Danielle Maddix, Hao Wang, and Yuyang Wang.
\newblock Domain adaptation for time series forecasting via attention sharing.
\newblock In {\em International Conference on Machine Learning}, pages 10280--10297. PMLR, 2022.

\bibitem{kaplan2020scaling}
Jared Kaplan, Sam McCandlish, Tom Henighan, Tom~B Brown, Benjamin Chess, Rewon Child, Scott Gray, Alec Radford, Jeffrey Wu, and Dario Amodei.
\newblock Scaling laws for neural language models.
\newblock {\em arXiv preprint arXiv:2001.08361}, 2020.

\bibitem{loshchilov2017decoupled}
Ilya Loshchilov and Frank Hutter.
\newblock Decoupled weight decay regularization.
\newblock {\em arXiv preprint arXiv:1711.05101}, 2017.

\bibitem{makansi2019overcoming}
Osama Makansi, Eddy Ilg, Ozgun Cicek, and Thomas Brox.
\newblock Overcoming limitations of mixture density networks: A sampling and fitting framework for multimodal future prediction.
\newblock In {\em Proceedings of the IEEE/CVF Conference on Computer Vision and Pattern Recognition}, pages 7144--7153, 2019.

\bibitem{mentzer2023finite}
Fabian Mentzer, David Minnen, Eirikur Agustsson, and Michael Tschannen.
\newblock Finite scalar quantization: Vq-vae made simple.
\newblock {\em arXiv preprint arXiv:2309.15505}, 2023.

\bibitem{montali2024waymo}
Nico Montali, John Lambert, Paul Mougin, Alex Kuefler, Nicholas Rhinehart, Michelle Li, Cole Gulino, Tristan Emrich, Zoey Yang, Shimon Whiteson, et~al.
\newblock The waymo open sim agents challenge.
\newblock {\em Advances in Neural Information Processing Systems}, 36, 2024.

\bibitem{nayakanti2023wayformer}
Nigamaa Nayakanti, Rami Al-Rfou, Aurick Zhou, Kratarth Goel, Khaled~S Refaat, and Benjamin Sapp.
\newblock Wayformer: Motion forecasting via simple \& efficient attention networks.
\newblock In {\em 2023 IEEE International Conference on Robotics and Automation (ICRA)}, pages 2980--2987. IEEE, 2023.

\bibitem{ngiam2021scene}
Jiquan Ngiam, Benjamin Caine, Vijay Vasudevan, Zhengdong Zhang, Hao-Tien~Lewis Chiang, Jeffrey Ling, Rebecca Roelofs, Alex Bewley, Chenxi Liu, Ashish Venugopal, et~al.
\newblock Scene transformer: A unified architecture for predicting multiple agent trajectories.
\newblock {\em arXiv preprint arXiv:2106.08417}, 2021.

\bibitem{orozco2020zero}
Bernardo~P{\'e}rez Orozco and Stephen~J Roberts.
\newblock Zero-shot and few-shot time series forecasting with ordinal regression recurrent neural networks.
\newblock {\em arXiv preprint arXiv:2003.12162}, 2020.

\bibitem{philion2023trajeglish}
Jonah Philion, Xue~Bin Peng, and Sanja Fidler.
\newblock Trajeglish: Learning the language of driving scenarios.
\newblock {\em arXiv preprint arXiv:2312.04535}, 2023.

\bibitem{radford2019language}
Alec Radford, Jeffrey Wu, Rewon Child, David Luan, Dario Amodei, Ilya Sutskever, et~al.
\newblock Language models are unsupervised multitask learners.
\newblock {\em OpenAI blog}, 1(8):9, 2019.

\bibitem{rawte2023survey}
Vipula Rawte, Amit Sheth, and Amitava Das.
\newblock A survey of hallucination in large foundation models, 2023.

\bibitem{salzmann2020trajectron++}
Tim Salzmann, Boris Ivanovic, Punarjay Chakravarty, and Marco Pavone.
\newblock Trajectron++: Dynamically-feasible trajectory forecasting with heterogeneous data.
\newblock In {\em Computer Vision--ECCV 2020: 16th European Conference, Glasgow, UK, August 23--28, 2020, Proceedings, Part XVIII 16}, pages 683--700. Springer, 2020.

\bibitem{seff2023motionlm}
Ari Seff, Brian Cera, Dian Chen, Mason Ng, Aurick Zhou, Nigamaa Nayakanti, Khaled~S Refaat, Rami Al-Rfou, and Benjamin Sapp.
\newblock Motionlm: Multi-agent motion forecasting as language modeling.
\newblock In {\em Proceedings of the IEEE/CVF International Conference on Computer Vision}, pages 8579--8590, 2023.

\bibitem{shi2022motion}
Shaoshuai Shi, Li~Jiang, Dengxin Dai, and Bernt Schiele.
\newblock Motion transformer with global intention localization and local movement refinement.
\newblock {\em Advances in Neural Information Processing Systems}, 35:6531--6543, 2022.

\bibitem{shi2023mtr++}
Shaoshuai Shi, Li~Jiang, Dengxin Dai, and Bernt Schiele.
\newblock Mtr++: Multi-agent motion prediction with symmetric scene modeling and guided intention querying.
\newblock {\em arXiv preprint arXiv:2306.17770}, 2023.

\bibitem{sima2023drivelm}
Chonghao Sima, Katrin Renz, Kashyap Chitta, Li~Chen, Hanxue Zhang, Chengen Xie, Ping Luo, Andreas Geiger, and Hongyang Li.
\newblock Drivelm: Driving with graph visual question answering.
\newblock {\em arXiv preprint arXiv:2312.14150}, 2023.

\bibitem{song2020pip}
Haoran Song, Wenchao Ding, Yuxuan Chen, Shaojie Shen, Michael~Yu Wang, and Qifeng Chen.
\newblock Pip: Planning-informed trajectory prediction for autonomous driving.
\newblock In {\em Computer Vision--ECCV 2020: 16th European Conference, Glasgow, UK, August 23--28, 2020, Proceedings, Part XXI 16}, pages 598--614. Springer, 2020.

\bibitem{suo2021trafficsim}
Simon Suo, Sebastian Regalado, Sergio Casas, and Raquel Urtasun.
\newblock Trafficsim: Learning to simulate realistic multi-agent behaviors.
\newblock In {\em Proceedings of the IEEE/CVF Conference on Computer Vision and Pattern Recognition}, pages 10400--10409, 2021.

\bibitem{tian2024drivevlm}
Xiaoyu Tian, Junru Gu, Bailin Li, Yicheng Liu, Chenxu Hu, Yang Wang, Kun Zhan, Peng Jia, Xianpeng Lang, and Hang Zhao.
\newblock Drivevlm: The convergence of autonomous driving and large vision-language models.
\newblock {\em arXiv preprint arXiv:2402.12289}, 2024.

\bibitem{torgo1996regression}
Lu{\'\i}s Torgo and Jo{\~a}o Gama.
\newblock Regression by classification.
\newblock In {\em Advances in Artificial Intelligence: 13th Brazilian Symposium on Artificial Intelligence, SBIA'96 Curitiba, Brazil, October 23--25, 1996 Proceedings 13}, pages 51--60. Springer, 1996.

\bibitem{touvron2023llama}
Hugo Touvron, Thibaut Lavril, Gautier Izacard, Xavier Martinet, Marie-Anne Lachaux, Timoth{\'e}e Lacroix, Baptiste Rozi{\`e}re, Naman Goyal, Eric Hambro, Faisal Azhar, et~al.
\newblock Llama: Open and efficient foundation language models.
\newblock {\em arXiv preprint arXiv:2302.13971}, 2023.

\bibitem{touvron2023llama2}
Hugo Touvron, Louis Martin, Kevin Stone, Peter Albert, Amjad Almahairi, Yasmine Babaei, Nikolay Bashlykov, Soumya Batra, Prajjwal Bhargava, Shruti Bhosale, et~al.
\newblock Llama 2: Open foundation and fine-tuned chat models.
\newblock {\em arXiv preprint arXiv:2307.09288}, 2023.

\bibitem{van2017neural}
Aaron Van Den~Oord, Oriol Vinyals, et~al.
\newblock Neural discrete representation learning.
\newblock {\em Advances in neural information processing systems}, 30, 2017.

\bibitem{varadarajan2022multipath++}
Balakrishnan Varadarajan, Ahmed Hefny, Avikalp Srivastava, Khaled~S Refaat, Nigamaa Nayakanti, Andre Cornman, Kan Chen, Bertrand Douillard, Chi~Pang Lam, Dragomir Anguelov, et~al.
\newblock Multipath++: Efficient information fusion and trajectory aggregation for behavior prediction.
\newblock In {\em 2022 International Conference on Robotics and Automation (ICRA)}, pages 7814--7821. IEEE, 2022.

\bibitem{wang2023multiverse}
Yu~Wang, Tiebiao Zhao, and Fan Yi.
\newblock Multiverse transformer: 1st place solution for waymo open sim agents challenge 2023.
\newblock {\em arXiv preprint arXiv:2306.11868}, 2023.

\bibitem{wilson2023argoverse}
Benjamin Wilson, William Qi, Tanmay Agarwal, John Lambert, Jagjeet Singh, Siddhesh Khandelwal, Bowen Pan, Ratnesh Kumar, Andrew Hartnett, Jhony~Kaesemodel Pontes, et~al.
\newblock Argoverse 2: Next generation datasets for self-driving perception and forecasting.
\newblock {\em arXiv preprint arXiv:2301.00493}, 2023.

\bibitem{yu2022scaling}
Jiahui Yu, Yuanzhong Xu, Jing~Yu Koh, Thang Luong, Gunjan Baid, Zirui Wang, Vijay Vasudevan, Alexander Ku, Yinfei Yang, Burcu~Karagol Ayan, et~al.
\newblock Scaling autoregressive models for content-rich text-to-image generation.
\newblock {\em arXiv preprint arXiv:2206.10789}, 2(3):5, 2022.

\bibitem{yu2024language}
Lijun Yu, José Lezama, Nitesh~B. Gundavarapu, Luca Versari, Kihyuk Sohn, David Minnen, Yong Cheng, Vighnesh Birodkar, Agrim Gupta, Xiuye Gu, Alexander~G. Hauptmann, Boqing Gong, Ming-Hsuan Yang, Irfan Essa, David~A. Ross, and Lu~Jiang.
\newblock Language model beats diffusion -- tokenizer is key to visual generation, 2024.

\bibitem{zhong2023guided}
Ziyuan Zhong, Davis Rempe, Danfei Xu, Yuxiao Chen, Sushant Veer, Tong Che, Baishakhi Ray, and Marco Pavone.
\newblock Guided conditional diffusion for controllable traffic simulation.
\newblock In {\em 2023 IEEE International Conference on Robotics and Automation (ICRA)}, pages 3560--3566. IEEE, 2023.

\bibitem{9636795}
Jinyun Zhou, Rui Wang, Xu~Liu, Yifei Jiang, Shu Jiang, Jiaming Tao, Jinghao Miao, and Shiyu Song.
\newblock Exploring imitation learning for autonomous driving with feedback synthesizer and differentiable rasterization.
\newblock In {\em 2021 IEEE/RSJ International Conference on Intelligent Robots and Systems (IROS)}, pages 1450--1457, 2021.

\bibitem{zhou2023query}
Zikang Zhou, Jianping Wang, Yung-Hui Li, and Yu-Kai Huang.
\newblock Query-centric trajectory prediction.
\newblock In {\em Proceedings of the IEEE/CVF Conference on Computer Vision and Pattern Recognition}, pages 17863--17873, 2023.

\bibitem{zhou2023qcnext}
Zikang Zhou, Zihao Wen, Jianping Wang, Yung-Hui Li, and Yu-Kai Huang.
\newblock Qcnext: A next-generation framework for joint multi-agent trajectory prediction.
\newblock {\em arXiv preprint arXiv:2306.10508}, 2023.

\end{thebibliography}
\newpage{}
\appendix

\section{Appendix}
\subsection{Implementation and Simulation Inference} \label{sec:5:2}

\paragraph{Architecture details}
Table~\ref{tab:6} summarizes the hyperparameters of the different models used in our implementation. We train a single model to generate the future motion of all three categories (i.e., Vehicle, Pedestrian, Cyclist), with each category having its own motion token vocabulary. The input road token feature contains three types of information: the position of each road token point, the road token direction at each point, and the type of each road token. For the prediction head in each decoder layer, we use a three-layer MLP, and the model weights are not shared across different decoder layers.

\begin{table}[htb]
\caption{Hyperparameters of different SMART models}
\centering
\scalebox{0.8}{
\setlength{\tabcolsep}{3pt}
\begin{tabular}{cccccc} 
\hline
\multirow{2}{*}{Modules} & \multirow{2}{*}{Hyperparameters}  & \multicolumn{4}{c}{Values} \\ 
\cline{3-6}
 &   & SMART 1M & SMART 7M & SMART 26M & SMART 101M \\ 
\hline
RoadNet & Number of self attention layers  & 1 & 3& 1& 3\\
 & Road token embeddings  & 64 & 128 & 256& 512\\
 & Size of road token vocabulary  & 1024 & 1024 & 1024 & 1024\\
 & Road token attention radius  & 10 & 10 & 10 & 10 \\
 \hline
MotionNet & Number of temporal attention layers  & 1 & 6& 6 & 6 \\
 & Number of agent-agent attention layers  & 2 & 6& 6 & 6 \\
 & Number of map-agent attention layers  & 2 & 6& 6 & 6 \\
 & Number of attention head  & 8 &  8&  8& 8 \\
 & Dimension of attention head  & 8 & 16 &  32& 64 \\
 & Feature dimension of Agent token embeddings  & 64 &  128&  256&512  \\
 & Size of motion token vocabulary  & 512 & 1024 & 1024 & 2048 \\
 \hline
 SMART& Total parameters  & 1.0M & 7.2M &26.9M  & 101.0M \\
\hline
\end{tabular}
}
\label{tab:6}
\end{table}

\paragraph{Training details}
The simulation model is trained end-to-end for all three agent types using the AdamW optimizer \cite{loshchilov2017decoupled}. Both the dropout rate and the weight decay rate are set to 0.1. The learning rate is decayed from 0.0002 to 0 using a cosine annealing scheduler. Training includes all vehicles within a scene. The batch size is set to 4, with a maximum GPU memory usage of 30GB.

\paragraph{Inference for WOSAC}
The test set comprises 44,920 scenes, and each scene requires running the model inference $32\times T$ times to generate the 32 simulations for a group of agents. During model inference, each simulation step produces the classified distribution of next tokens.  There are two options for next token sampling: selecting the maximum-likelihood token or sampling among the top-k motion tokens with the redistributed probability. The first approach, while accurate, tends to yield less varied generations. Conversely, opting for the top-k motion tokens encourages diversity but can compound errors, generating trajectories with unrealistic kinematic motions or even drift. To balance realism and diversity, we use top-5 sampling at every step during the simulation. Videos of rollouts can be found on our project page or supplementary materials. For each scenario, the SMART model directly controls all agents within the scene. Due to the focus of this article on the generalization and scalability of the model, we have achieved good results in specific scene generation without extensive exploration of detailed tricks.

\subsection{Detailed comparison in the WOSAC leaderboard} \label{sec:5:1}
\begin{table}[htb]
\caption{Per-component metric results on the test split of WOMD, representing likelihoods. Due to updates in the calculation of WOSC evaluation metrics, methods are ranked by the composite metric on the \href{https://waymo.com/open/challenges/2023/sim-agents/}{2023 Leaderboard} for a broader comparison. For latest WOSC, please refer directly to the updated \href{https://waymo.com/open/challenges/2024/sim-agents/}{2024 Leaderboard} for detailed comparisons.}
\centering
\scalebox{0.65}{
\begin{tabular}{ccccccccccc} 
\hline
\multirow{2}{*}{Method} & \multicolumn{4}{c}{KINEMATIC} & \multicolumn{3}{c}{INTERACTIVE} & \multicolumn{2}{c}{MAP} & \multirow{2}{*}{minADE$\downarrow$} \\ 
\cline{2-10}
 & \begin{tabular}[c]{@{}c@{}}LINEAL \\SPEED$\uparrow$\end{tabular} & \begin{tabular}[c]{@{}c@{}}LINEAR \\ACCEL$\uparrow$\end{tabular} & \begin{tabular}[c]{@{}c@{}}ANG. \\SPEED$\uparrow$\end{tabular} & \begin{tabular}[c]{@{}c@{}}ANG. \\ACCEL$\uparrow$~\end{tabular} & \begin{tabular}[c]{@{}c@{}}DIST TO \\OBJ.$\uparrow$\end{tabular} & COLLISION $\uparrow$& TTC $\uparrow$& \begin{tabular}[c]{@{}c@{}}DIST TO \\ROAD$\uparrow$\end{tabular} & \begin{tabular}[c]{@{}c@{}}OFF\\~ROAD$\uparrow$\end{tabular} &  \\ 
\hline
WAYFORMER & 0.202 & 0.144 & 0.248 & 0.312 & 0.192 & 0.449 & 0.766 & 0.379 & 0.305 & 6.823 \\
SBTA-ADIA & 0.317 & 0.174 & 0.478 & 0.463 & 0.265 & 0.337 & 0.770 & 0.557 & 0.483 & 3.611 \\
CAD & 0.346 & 0.252 & 0.432 & 0.311 & 0.33 & 0.311 & 0.789 & 0.637 & 0.539 & 2.314 \\
JOINT-MULTIPATH++ & 0.431 & 0.230 & 0.019 & 0.035 & 0.349 & 0.485 & 0.811 & 0.637 & 0.613 & 2.051 \\
MTR+++ & 0.411 & 0.106 & 0.483 & 0.436 & 0.345 & 0.414 & 0.796 & 0.654 & 0.577 & 1.681 \\
QCNeXt & \textbf{0.477} & 0.242 & 0.325 & 0.198 & 0.375 & 0.324 & 0.756 & 0.609 & 0.360 & \textbf{1.083} \\
MVTE & 0.442 & 0.221 & 0.535 & 0.481 & 0.382 & 0.450 & 0.832 & 0.664 & 0.640 & 1.677 \\
Trajeglish & 0.450 & 0.192 & \textbf{0.538} & 0.485 & \textbf{0.387} & 0.922 & \textbf{0.836} & \textbf{0.659} & 0.886 & 1.571 \\
SMART 7M & 0.363 & \textbf{0.296} & 0.423 & \textbf{0.564} & 0.376 & \textbf{0.963} & 0.832 & \textbf{0.659} & 0\textbf{.936} & 1.749 \\
\hline
\end{tabular}
}
\label{tab:5}
\end{table}

The Waymo Open Sim Agents Challenge (WOSAC) is a significant initiative aimed at advancing the development and evaluation of simulation agents for autonomous vehicles. This challenge leverages the Waymo Open Motion Dataset (WOMD) to provide high-fidelity object behaviors and shapes produced by a state-of-the-art offboard perception system. Participants are required to simulate scenarios involving up to 128 agents, focusing on closed-loop evaluation to ensure realism in agent behaviors and interactions. The evaluation framework employs various metrics, including kinematic features, interaction-based features, and map-based features, to assess the performance of simulation agents in generating realistic and diverse behaviors that match real-world driving data. WOSAC computes three metrics over nine measurements: kinematic metrics (linear speed, linear acceleration, angular speed, angular acceleration magnitude), object interaction metrics (distance to nearest object, collisions, time-to-collision), and map-based metrics (distance to road edge, road departures).

In the benchmark comparisons presented in Table \ref{tab:5}, the SMART 7M method, developed by our team, demonstrates superior performance across multiple metrics, particularly excelling in interactive and safety-related indicators. Notably, SMART 7M achieved the highest scores in angular acceleration, distance to nearest object, collision avoidance, and off-road metrics, underscoring its effectiveness in complex driving scenarios. These results highlight the robustness of SMART 7M in ensuring safety and reliability, indicating its advanced capability in managing dynamic and potentially hazardous traffic conditions more effectively than other evaluated methods. This performance also suggests the potential of the SMART model to be applied to planning tasks.

\subsection{Additional ablation studies} \label{sec:5:3}
\paragraph{Comparison of the scalability and generalization of different architectures}
This section presents experiments comparing the architecture proposed in this paper with the MVTE model. The MVTE model\footnote{Since MVTE does not have open-source code, we reproduced the results by relying on the MTR model}, derived from MTR, represents continuous distribution regression models. 
\begin{figure}[htb]
\centering
\includegraphics[width=0.8\linewidth]{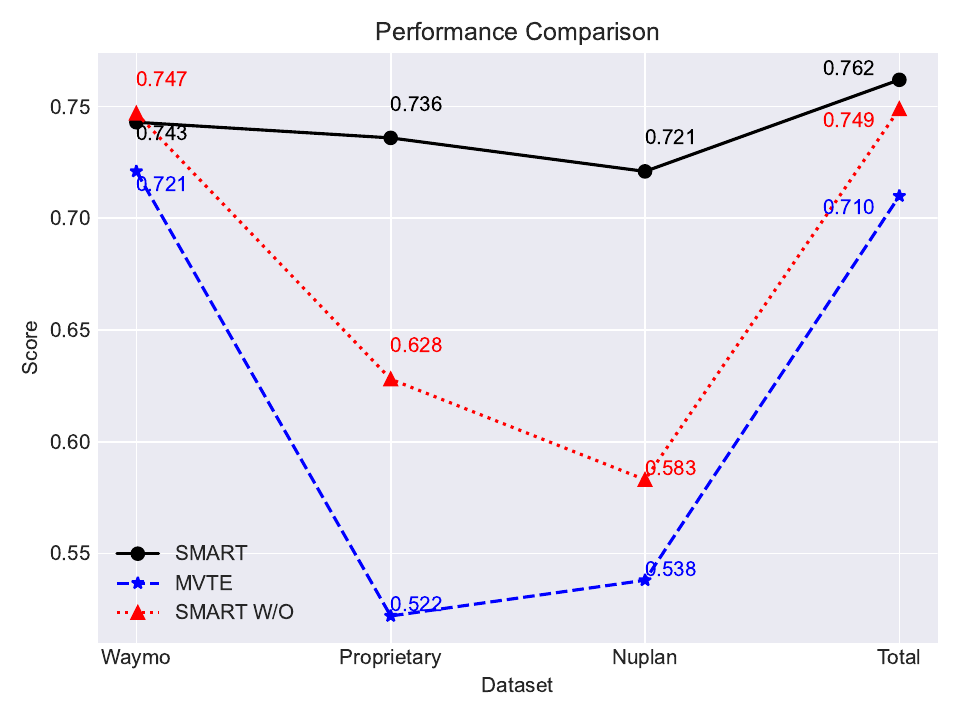}
\caption{SMART w/o refers to SMART model without the road vector tokenization and noise tricks proposed in this paper. To ensure the fairness of the experiments, all model parameters were adjusted to the 90-100M. Models were trained on various datasets and validated only on the WOMD validation dataset}
\label{fig:5}
\end{figure}
The experimental results shown in Figure~\ref{fig:5} indicate that distribution regression-based models have poor generalization capabilities across different datasets. Models trained with incremental data from other datasets performed worse overall than models trained solely on the WOMD. An interesting phenomenon is that although our private dataset contains more data than the NuPlan dataset, the performance of MVTE trained on it was inferior to that on the NuPlan dataset. This suggests that the distribution regression-based paradigm is likely to cause the model to overfit to a dataset. From the SMART w/o results, it can be seen that the model's generalization performance is poor, but the effect of incremental data can improve the performance compared to a single training dataset. Based on the above experiments, it can be inferred that discrete tokenization is a very effective way to eliminate the dataset gap. Moreover, autoregressive models based on cross-entropy classification loss are key to the scalability of trajectory generation models, which aligns with why large language models (LLMs) have significant scaling capabilities.

\paragraph{Comparison of different tokenizer}
In Trajeglish\cite{philion2023trajeglish}, a detailed comparison of various discretized tokenizers is conducted. As introduced in Sec.~\ref{sec:3.1} of this paper, we ultimately adopted the k-disks approach for token vocabulary construction. Prior to our work, no studies had attempted to construct a vocabulary for agent and road motion tokens using latent tokenizer methods \cite{van2017neural}. Therefore, we drew on the visual domain's VQ-VAE approach to perform latent autoencoding of motion tokens and provided a comparison of this tokenizer with the method selected in this paper.

\begin{table*}[h]
\caption{Comparison of different tokenizer. Experimental results are based on the SMART 7M}
\centering
\setlength{\tabcolsep}{2pt} 
\begin{tabular}{@{}ccccccc} 
\hline
    & \multicolumn{3}{c}{Train on WOMD}& \multicolumn{3}{c}{Train on NuPlan}\\
    \hline
    \begin{tabular}{c}Tokenizer\end{tabular}& \begin{tabular}{c}Kinematics\end{tabular}& \begin{tabular}{c}Interactive\end{tabular}& \begin{tabular}{c}Map\end{tabular}& \begin{tabular}{c}Kinematics\end{tabular}& \begin{tabular}{c}Interactive\end{tabular}&\begin{tabular}{c}Map\end{tabular}\\
    \hline
    VQ-VAE& 0.461& 0.810& 0.853& 0.376& 0.687&0.703\\
    K-disks& 0.453& 0.803& 0.851& 0.416&0.785& 0.797\\
\hline
\end{tabular}
\label{tab:7}
\end{table*}
From the results in Table~\ref{tab:7}, it is evident that VQ-VAE performs better on a single dataset compared to k-disks. Specifically, both methods achieve similar results in interactive and map-based metrics, but VQ-VAE outperforms k-disks in kinematic metrics. The k-disks approach loses fine-grained trajectory information during discretization, whereas VQ-VAE better fits the true distribution of the dataset when reconstructing trajectories. However, when comparing the two methods' performance in zero-shot generalization, k-disks significantly outperform VQ-VAE. We speculate that during the training of the VQ-VAE tokenizer to construct motion and road token vocabularies, the tokenizer may have already memorized or overfitted to the training dataset. Therefore, to achieve better generalization performance using the VQ-VAE approach, it is essential to pre-train the VQ-VAE tokenizer on a large-scale dataset.

\paragraph{Comparison of SMART models with different scales}
For language models, large and diverse datasets are relatively easy to obtain. In contrast, the autonomous driving motion domain lacks a data source of comparable size and diversity. To validate scaling laws on a larger dataset, we integrated data from Waymo, Nuplan, and our proprietary dataset. We introduced our proprietary dataset solely for validating scaling laws. For the WOSC leaderboard evaluation, we exclusively used the Waymo dataset. For generalization and other ablation experiments, we utilized both Nuplan and Waymo open-source datasets to facilitate reproducibility of the experiments by providing access to widely available datasets. Table \ref{tab:8} below summarizes the scenario count, duration, and total motion token count for each dataset.

\begin{table}
    \centering
     \caption{Data sources}
    \begin{tabular}{cccc}
    \toprule
         Dataset&  Scene Count&  Single Scenario Duration & Total Motion Token Count\\
    \midrule
         Nuplan&  30w&  10s& 0.13B\\
         Waymo&  48w&  9s& 0.18B\\
         Proprietary&  150w&  11s& 0.68B\\
        Total& 228w& - & 1B\\
    \bottomrule
    \end{tabular}
   
    \label{tab:8}
\end{table}

\begin{table}[ht]
    \caption{Comparison of SMART models with different scales. Training time refers to the duration required for the model to converge using the entire dataset. Inference time refers to the time taken by the model to predict the next token for a single frame.}
    \centering
    \scalebox{0.9}{
    \begin{tabular}{cccccc}
    \toprule
         Method&  \begin{tabular}{c}Kinematic \\ metrics$\uparrow$\end{tabular}&  \begin{tabular}{c} Interactive \\metrics$\uparrow$\end{tabular} &  \begin{tabular}{c} Map-based \\ metrics$\uparrow$\end{tabular} &   Training time&Average inference time\\
    \midrule
    SMART 1M& 0.423& 0.782& 0.835& 8hours&10.30ms\\
    SMART 7M & 0.436& 0.809& 0.852& 23hours&17.21ms\\
    SMART 26M & 0.442& 0.817& 0.864&   3days&25.94ms\\
    SMART 101M &  \textbf{0.457} &  \textbf{0.819} &  \textbf{0.872}&   1week&46.58ms\\
    \bottomrule
    \end{tabular}
    }
    \label{tab:9}
\end{table}

 The results in Table \ref{tab:9} highlight the performance of SMART models with different parameter scales across various metrics. As the model scale increases from SMART 1M to SMART 101M, there is a significant improvement in both the interactive metrics and the map-based metrics. This indicates that larger models are better at capturing interactions and understanding map-based context, leading to enhanced performance in these areas. However, the kinematic metrics show minimal variation. Additionally, the training time and average inference time increase substantially with larger models, reflecting the trade-off between model performance and computational cost. Validation is conducted every 50,000 train steps. The model is considered to have converged if there is no significant loss reduction or metric improvement after five consecutive validations. The training and inference time is measured on 32 NVIDIA TESLA V100 GPUs.



\end{document}